\documentclass[
]{ceurart}

\begin{document}

\copyrightyear{2023}
\copyrightclause{Copyright for this paper by its authors.
  Use permitted under Creative Commons License Attribution 4.0
  International (CC BY 4.0).}

\conference{De-Factify 2.0: Workshop on Multimodal Fact Checking and Hate Speech Detection, co-located with AAAI 2023. 2023 DC, USA}

\title{Team Triple-Check at Factify 2: Parameter-Efficient Large Foundation Models with Feature Representations for Multi-Modal Fact Verification}

\author[1]{Wei-Wei Du}[%
orcid=,
email=wwdu.cs10@nycu.edu.tw
]

\author[1]{Hong-Wei Wu}[%
orcid=,
email=johnnyhwu.cs11@nycu.edu.tw
]

\author[1]{Wei-Yao Wang}[%
orcid=,
email=sf1638.cs05@nctu.edu.tw
]

\address[1]{Department of Computer Science, National Yang Ming Chiao Tung University, Hsinchu, Taiwan}

\author[1]{Wen-Chih Peng}[%
email=wcpeng@cs.nycu.edu.tw
]

\begin{abstract}
Multi-modal fact verification has become an important but challenging issue on social media due to the mismatch between the text and images in the misinformation of news content, which has been addressed by considering cross-modalities to identify the veracity of the news in recent years.
In this paper, we propose the \textbf{Pre-CoFactv2} framework with new parameter-efficient foundation models for modeling fine-grained text and input embeddings with lightening parameters, multi-modal multi-type fusion for not only capturing relations for the same and different modalities but also for different types (i.e., claim and document), and feature representations for explicitly providing metadata for each sample.
In addition, we introduce a unified ensemble method to boost model performance by adjusting the importance of each trained model with not only the weights but also the powers.
Extensive experiments show that Pre-CoFactv2 outperforms Pre-CoFact by a large margin and achieved new state-of-the-art results at the Factify challenge at AAAI 2023.
We further illustrate model variations to verify the relative contributions of different components.
Our team won the first prize (F1-score: 81.82\%) and we made our code publicly available\footnote{https://github.com/wwweiwei/Pre-CoFactv2-AAAI-2023.}.
\end{abstract}


\maketitle

\section{Introduction}

The rapid rise of social media technology allows people to send and receive information immediately, and also creates fertile soil for the fast spread of fake news.
The proliferation of fake news not only triggers a storm of public opinion but also manipulates public events such as elections.
For instance, there were approximately 30 million tweets from 2.2 million users on Twitter in the five months preceding the US 2016 presidential election, and either fake or extremely biased news was contained in 25\% of these tweets \cite{DBLP:journals/corr/abs-1803-08491}, misleading people and seriously influencing the outcome of the election.
Moreover, this issue became even worse during the COVID-19 pandemic period.
Unconfirmed news with eye-catching images is becoming popular on social media since richer information easily attracts more viewers than news with only text.
Therefore, in order to mitigate the negative impact caused by fake news, there is an urgent need to develop a multi-modal fake checker that can automatically assess the validity of news.

Given a claim and the support information consisting of not only text but also images, the fact checker aims to discriminate whether the claim entails the support.
Recent approaches have demonstrated that utilizing multi-modal contexts to detect fake news achieves better performance than only using one modality.
For instance, \citet{DBLP:conf/kdd/WangMJYXJSG18} learned event-invariant features using an adversarial network along with a multi-modal feature extractor to enhance the performance of the fake news detector.
More recently, \citet{DBLP:conf/aaai/WangP22} introduced Pre-CoFact with DeBERTa \cite{DBLP:conf/iclr/HeLGC21} and DeiT \cite{DBLP:conf/icml/TouvronCDMSJ21}, to extract features from both claims and documents' text and images, respectively and then fuses multi-modal contexts with the co-attention modules, illustrating the competitive performance without auxiliary information.

However, Pre-CoFact fails to finetune the pre-trained model due to the large number of parameters, which heavily depend on the post layers to learn the contextual information.
Besides, we argue that additional textual features such as stopword and URL counts provide definitive descriptions and relations between inputs.
In addition, the co-attention modules only capture the dependencies between modalities while ignoring the relations between different types of samples (i.e., claim and document).
Therefore, we believe that multi-modal fact verification remains an unexplored but essential problem.

\begin{figure*}
  \centering
  \includegraphics[width=\linewidth]{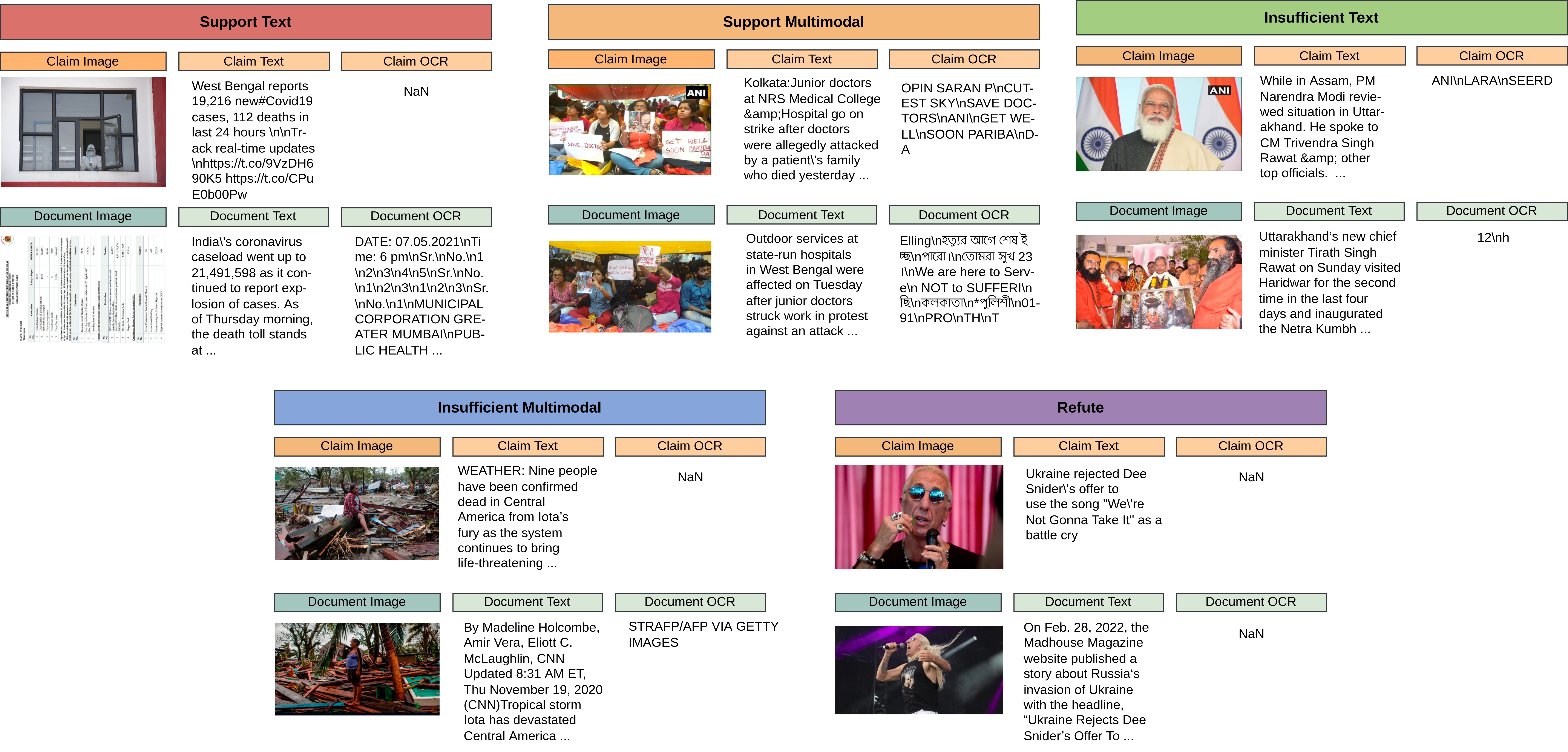}
  \caption{An example of each category in the Factify 2 dataset \cite{surya2023factify2}.
  Each sample contains a claim and document, both of which are composed of an image, text, and OCR of the image.}
  \label{fig:dataset-sample}
\end{figure*}

To tackle this task and the aforementioned limitations, we examine our proposed method on the real-world Factify 2 challenge \cite{surya2023factify2}, which is the largest multi-modal fact verification dataset consisting of 50K news items from India and the US.
Specifically, each sample contains one \textit{document} with a related image representing the reliable source of information and one \textit{claim} which also includes an associated image representing another source of information whose validity needs to be assessed.
The training set, validation set, and testing set are composed of 35000 samples, 7500 samples, and 7500 samples, respectively.
As shown in Figure \ref{fig:dataset-sample}, each sample contains a textual claim, claim image, optical character recognition (OCR) of the claim image, document, document image, and OCR of a document image, and is classified into support, insufficient evidence, and refute between given claims and documents (detailed labels will be introduced in $\S$\ref{problem})

In this paper, we propose \textbf{Pre-CoFactv2} with parameter-efficient large foundation models with feature representations to address the challenge.
We utilize large-scale pre-trained foundation models, DeBERTa \cite{DBLP:conf/iclr/HeLGC21} and Swin Transformer v2 (Swinv2) \cite{DBLP:conf/cvpr/Liu0LYXWN000WG22}, to extract contextualized embeddings from both textual content and visual images, respectively, and an adapter module is used to improve model performance by finetuning the backbone with only a few parameters.
Furthermore, additional feature metadata such as the number of stopwords and URLs in the claim and document is generated from the input to enrich the information; this enables the model to learn explicit information from different perspectives.
Afterwards, we integrate the information among several modalities with multi-modal multi-type fusion modules into corresponding embeddings to classify the category of the news.

To summarize, the contributions of this paper are three-fold:

\begin{itemize}
    \item We propose parameter-efficient large foundation models with feature representations (Pre-CoFactv2) for multi-modal fact verification by integrating adapters in the large-scale foundation models to achieve competitive performance by training only lightening parameters, and converting additional features to learn the explicit correlations between claim and document.
    \item In addition to capturing information between modalities, we design multi-modal multi-type fusions for different types of modalities (i.e., claim text and document images, document text and claim images), which enforces the model to distinguish between the given claim and document.
    \item To boost the detection quality, we introduce a unified ensemble method to integrate various considerations from diverse models. This approach won first place, surpassing the second place by 1.3\% and the official baseline by 25.9\% in terms of testing score. Moreover, extensive experiments were conducted to examine and analyze the contribution and effectiveness of each module.
\end{itemize}
\section{Related Works}
\subsection{Multi-Modal Fact Verification}

In recent years, multiple modalities (e.g., text and images) have been incorporated to demonstrate the great potential for fact verification.
MAVE \cite{DBLP:conf/www/KhattarG0V19} was proposed to learn better multi-modal shared representations with a variational autoencoder by jointly training with a fact verification classifier to verify the posts. 
In addition to learning shared features among several modalities, \citet{DBLP:conf/sigir/QianWHFX21} introduced HMCAN with a multi-modal contextual attention module to model the features from multiple modalities in each news post, and a hierarchical encoding module to capture the rich hierarchical semantics of text. 
Besides, \citet{DBLP:conf/acl/WuZZWX21} proposed MCAN to take inter-modality relations into consideration by using multiple co-attention layers to fuse visual and textual features extracted from Transformer-based models. 
Recently, \citet{DBLP:conf/www/0003LZSLTS22} proposed CAFE, which consists of an alignment module to transform the features from each modality into a shared semantic space, an ambiguity module to estimate the ambiguity between different modalities and a fusion module to capture the multi-modal correlations. 

Existing work focuses on the importance of learning shared representations among multi-modal information, which motivates us to use attention mechanisms to achieve the purpose.
However, pre-trained models with a base size are often used in the previous approaches, which cannot utilize more fine-grained features compared with the large-size model.
To that end, we adopt large-size pre-trained models to effectively produce embeddings of text and images from the complex multi-modal inputs.

\subsection{Pre-Trained Model for Different Modalities}

Transformer \cite{DBLP:conf/nips/VaswaniSPUJGKP17} has become the widely-used neural network architecture in various NLP and CV tasks due to its parallel computation and long-term considerations.
Since the advent of BERT \cite{DBLP:conf/naacl/DevlinCLT19}, a line of large-scale Transformer-based pre-trained language models (PLMs) such as GPT-3 \cite{DBLP:conf/nips/BrownMRSKDNSSAA20}, DeBERTa \cite{DBLP:conf/iclr/HeLGC21}, PaLM \cite{DBLP:journals/corr/abs-2204-02311}, and BLOOM \cite{DBLP:journals/corr/abs-2211-05100} have been introduced to demonstrate the generalizability of large-scale PLMs.
With the drastically increased capacity of the model, the accuracy of various language benchmarks has been significantly improved; it was therefore been adopted to finetune downstream tasks and has achieved better performance. 

In addition to the success of PLMs in NLP tasks, Transformer has also started taking over visual benchmarks recently.
\citet{DBLP:conf/iclr/DosovitskiyB0WZ21} proposed Vision Transformer (ViT) for pre-training with image patches, which has achieved competitive results compared to state-of-the-art convolution neural networks on ImageNet-1K image-level classification benchmarks.
Swin Transformer (Swinv1) \cite{DBLP:conf/iccv/LiuL00W0LG21} constructs hierarchical feature maps and uses the shifted window approach for computing self-attention, making it suitable as a general-purpose backbone for various vision tasks such as object detection and semantic segmentation.
Afterwards, Swin Transformer v2 (Swinv2) \cite{DBLP:conf/cvpr/Liu0LYXWN000WG22} was proposed with several adaptations in order to better scale up model capacity and window resolution to mitigate the unstable training and size discrepancy between the pre-training and training images of Swinv1.

To use the generic knowledge of textual and visual information for fact verification, we employ SOTA large-scale PLMs as pre-trained models instead of learning from scratch.
Furthermore, we utilize adapters in large-scale PLMs to finetune PLMs with lightening parameters while improving model performance.
\section{Method}
\subsection{Problem Formulation}
\label{problem}
Given a multi-modal claim denoted by $C = \{C^T_i, C^I_i\}^{|C|}_{i=1}$ and a fact-checking document denoted by $D = \{D^T_i, D^I_i\}^{|D|}_{i=1}$, the goal is to classify one of the five categories:
\begin{itemize}
    \item Support\_Text: the claim text is similar but images of the document and claim are not similar.
    \item Support\_Multimodal: both the claim text and image are similar to that of the document.
    \item Insufficient\_Text: both text and images of the claim are neither supported nor refuted by the document.
    \item Insufficient\_Multimodal: the claim text is neither supported nor refuted by the document but images are similar to the document.
    \item Refute: The images or text from the claim and document are completely contradictory.
\end{itemize}

Each sample contains a claim and a document, each of which includes the OCR feature, some special token count (e.g., stopword, @ URL), and word and character length.


\begin{figure*}
  \centering
  \includegraphics[width=\linewidth]{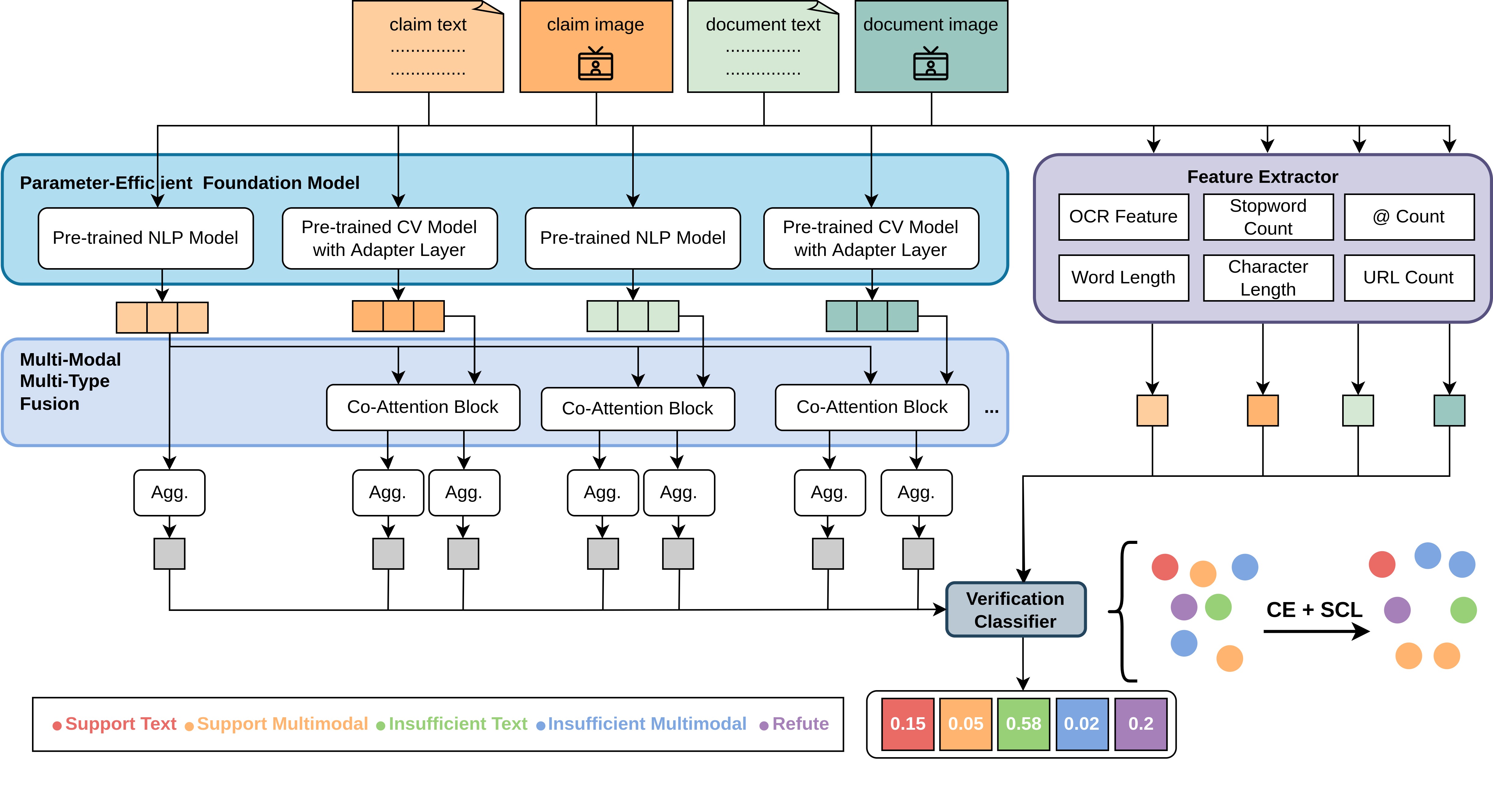}
  \caption{Illustration of the Pre-CoFactv2 framework.
  The parameter-efficient foundation model aims to transform the input text and images into embedding by the pre-trained language model. Then, the multi-modal multi-type fusion fuses the information from the same modality (images/text from the claim and document), different modalities (images and text from the claim/document), and different types (image from the claim and text from the document, and text from the claim and image from the document) to obtain contexts.
  Besides, the feature extractor is designed to convert input text and images into several features.
  In the end, the verification classifier contains cross-entropy loss and contrastive loss to predict the possible class based on the embeddings from previous outputs.}
  \label{fig:framework-overview}
\end{figure*}

\subsection{Pre-CoFactv2 Overview}
Figure \ref{fig:framework-overview} illustrates an overview of the proposed Pre-CoFactv2 framework.
The additional features are generated by the feature extractor from the given claim text, claim image, document text, and document image.
Then, we adopt two parameter-efficient foundation models for learning in-domain knowledge from pre-trained embeddings with adapters and a multi-modal multi-type fusion module for modeling not only cross-modality (i.e., text and image) relations but also cross-type (i.e., claim and document) relations.
Outputs of these embeddings are fused by the verification classifier with cross-entropy loss as well as supervised contrastive loss \cite{DBLP:conf/iclr/GunelDCS21} to separate embeddings and find clearer boundaries.

\subsection{Feature Extractor}
Inspired by previous works for textual information in fake checking \cite{DBLP:conf/www/CastilloMP11, DBLP:journals/corr/abs-2112-09253}, textual features are extracted from both claim and document text to enable the model to learn explicit information from different perspectives.
Specifically, we extract statistical features from the text, namely sentence word length, and character length.
Then, we calculate the stopword, @name, and URL counts to indicate the style of text content.
Besides, OCR text extracted from images is adopted to get the semantics information of the image instead of pixel values only.

\subsection{Parameter-Efficient Foundation Model}
\subsubsection{Foundation Model}
Benefiting from the advancement of utilizing basic knowledge as pre-trained models, we follow \cite{DBLP:conf/aaai/WangP22} to incorporate NLP and CV foundation models as the initialized embeddings of text and images.
However, previous work failed to fully exploit the knowledge of generic datasets; we, therefore, adopt state-of-the-art foundation models with larger sizes as the pre-trained models.
Specifically, we first use DeBERTa large \cite{DBLP:conf/iclr/HeLGC21} as our pre-trained NLP model and Swinv2 base \cite{DBLP:conf/cvpr/Liu0LYXWN000WG22} as our pre-trained CV model, and then the embedding layer is used for transforming pre-trained embeddings to embeddings in our task.
Formally, the $i$-th output of the embedding layer is calculated as follows:
\begin{equation}
    E_{C^I_i} = \sigma(W_{C^I}X_{C^I_i} + b_{C^I}); X_{C^I_i} = Swinv2(C^I_i),
\end{equation}
\begin{equation}
    E_{D^I_i} = \sigma(W_{D^I}X_{D^I_i} + b_{D^I}); X_{D^I_i} = Swinv2(D^I_i),
\end{equation}
\begin{equation}
    E_{C^T_i} = \sigma(W_{C^T}X_{C^T_i} + b_{C^T}); X_{C^T_i} = DeBERTa(C^T_i),
\end{equation}
\begin{equation}
    E_{D^T_i} = \sigma(W_{D^T}X_{D^T_i} + b_{D^T}); X_{D^T_i} = DeBERTa(D^T_i),
\end{equation}
where the dimensions of $E_{C^I_i}, E_{D^I_i}, E_{C^T_i}, E_{D^T_i}$ are $d$, the activation function $\sigma$ uses ReLU \cite{DBLP:journals/corr/abs-1803-08375}.

\subsubsection{Parameter-Efficient Adapter}
Finetuning existing foundation models requires a large number of computation resources due to the number of parameters, but is able to learn better representations for downstream tasks which are not trained in Pre-CoFact.
To that end, we design an adapter module in the foundation models, which enables us to finetune the backbone with only a few parameters but still achieves better performance than the freeze parameters.

We follow \cite{DBLP:conf/naacl/FuCLL22}, which demonstrates the competitive performance of fine-tuning between lightening parameters and all backbone parameters, by adding an additional adapter layer in the output layer of Swinv2 to empower the model to capture the information of images in the downstream tasks (fake news detection in this paper).
The weights of the adapter layer are computed as follows:
\begin{equation}
\begin{aligned}
    Swinv2(X) = FFN(\tilde{X}) + Adapter(\tilde{X});
    Adapter(\tilde{X}) = (W\tilde{X} + b) + v,
\end{aligned}
\end{equation}
where $\tilde{X}$ is the output embedding before the feed-forward layer of Swin-Transformerv2, $FFN(X)$ is the original feed-forward layer, and $Adapter(X)$ produces the adapted representations with the same dimension of $FFN(X)$ from the input-related weights.

We note that we cannot add adapter layers to our NLP foundation models since our GPU memory will OOM, but we believe that this concept can improve the performance as well.

\subsection{Multi-Modal Multi-Type Fusion}
Previous work only considered 1) images of claims and documents, 2) text of claims and documents, 3) images of claims and text of documents, and 4) images of claims and text of claims to produce context embeddings.
However, the relations between different types are critical to judge the text and images across claims and documents.
Therefore, we add two additional co-attention blocks to model correlations between 5) images of documents and text of claims, and 6) images of documents and text of documents.
The co-attention block is a variation of the multi-head self-attention block, which takes two modalities as inputs to learn interactions and relations.
Specifically, we compute the dot products of the query and the key, divide each by $\sqrt[]{d}$, and apply a softmax function to the attention scores to obtain the weights on the values, which indicates the relative importance of each value for a given query.

We illustrate the computation of images of claims ($E_{C^I}$) and documents ($E_{D^I}$) as examples, and the others follow a similar process:
\begin{equation}
    Q_C^{I} = E_{C^I} W^{Q_C^{I}}, K_C^{I} = E_{C^I} W^{K_C^{I}}, V_C^{I} = E_{C^I} W^{V_C^{I}},
\end{equation}
\begin{equation}
    Q_D^{I} = E_{D^I} W^{Q_D^{I}}, K_D^{I} = E_{D^I} W^{K_D^{I}}, V_D^{I} = E_{D^I} W^{V_D^{I}}
\end{equation}
\begin{equation}
    Att(E_{C^I}, E_{D^I}) = softmax(\frac{Q_C^I (K_D^{I})^T}{\sqrt{d}})V_D^I)
\end{equation}
\begin{equation}
    Z^I = Norm(E_{C^I} + Att(E_{C^I}, E_{D^I}))
\end{equation}
\begin{equation}
    O_{C, D}^I = Norm(FFN(Z^I)+Z^I)
\end{equation}
where $W^{Q_C}, W^{K_C}, W^{V_C}, W^{Q_D}, W^{K_D}, W^{V_D} \in \mathbb{R}^{d \times d}$, and $Norm$ and $FFN$ is the same normalization method and feed-forward network as in \cite{DBLP:conf/nips/VaswaniSPUJGKP17}.

To use fewer parameters, we share weights in a co-attention block for improving performance because of allowing the model to learn common representations of the inputs.
Finally, we apply the mean aggregation to fuse all the results into one embedding to represent the corresponding sentence or image embeddings.

\subsection{Category Classifier}
For predicting the label of the given claims and documents, all of the 12 aggregated outputs from the multi-modal multi-type fusion, the 4 aggregated embeddings of $E_{C^I_i}, E_{D^I_i}, E_{C^T_i}, E_{D^T_i}$, and the 32 dimensions output from the feature extractor are concatenated as the input $O$ of the classifier:

\begin{equation}
    \hat{y_i} = softmax((\sigma(O W^{Z1})) W^{Z2}),
\end{equation}
where $W^{Z1} \in \mathbb{R}^{48d \times d_m}$ and $W^{Z2} \in \mathbb{R}^{d_m \times 5}$.
Note that $\sigma$ uses ReLU which is the same as $E$.

To enhance the generalization of our method, we jointly train supervised contrastive learning \cite{DBLP:conf/iclr/GunelDCS21} and cross-entropy.
Thus, embeddings with the same label would become closer, while embeddings with different labels would increase the distance.
The loss function is as follows:
\begin{equation}
    \mathbb{L} = \alpha \times -\sum_{i=1}^{|C|} y_i log(\hat{y}_i) + (1-\alpha) \times SupConLoss,
\end{equation}
where $SupConLoss$ indicates a supervised contrastive loss.
We set 0.7 for cross-entropy loss and 0.3 for supervised contrastive loss respectively\footnote{We empirically found that supervised contrastive loss is not beneficial in this task ($\S$\ref{variations}), thus the final model sets $\alpha$ as 1.}.

\subsection{Unified Ensemble Techniques}
To eliminate the effect of noisy data and to integrate various advantages, ensemble learning with power weighted sum has been used to integrate the informative knowledge from different models to achieve a better predictive performance of the overall model via the voting technique \cite{DBLP:conf/aaai/WangP22,DBLP:conf/ltedi/WangTDP22}.
However, we argue that using different powers for each model improves the generalizability since each model does not require to use of the same projection space.
Therefore, we propose a unified ensemble method that the final predicted
probabilities $P$ are computed with the independent power weighted sum as: 
\begin{equation}
    P = P_1^{N_1} \times w_1 + \cdots + P_M^{N_M} \times w_M,
\end{equation}
where $M$ is set to 3 in this work, $w1, \cdots, w_M$ are weights of the corresponding model, and $N_1, \cdots, N_M$ are weights of power.
We tune these hyper-parameters based on the validation set and use ensemble weights as 0.2, 0.7, and 0.6 and powers as 0.125, 0.125, and 0.25, respectively.
\section{Results and Analysis}

\subsection{Implementation Details}
The dimension of additional features from text and image was set to 32, the embedding dimension $d$ of text and image were both set to 256, the inner dimension of the feed-forward layer was 512, and the number of heads was set to 12.
The hidden dimension of the classifier $d_m$ was 128.
The dropout rate was 0.1, and the max sequence length was 512.
The batch size was 24, the learning rates were set to 5e-5 and 1e-5, the pre-training epochs were set to 10 and the training epochs were set to 15, and the seeds were tested with 42.
The pre-trained DeBERTa was deberta-large\footnote{https://huggingface.co/microsoft/deberta-large}, and the Swinv2 was swinv2-base-patch4-window8-256\footnote{https://huggingface.co/microsoft/swinv2-base-patch4-window8-256}.
All of the parameters in the two pre-trained models are first finetuned by another dataset \cite{DBLP:conf/aaai/MishraSBCRPD0SE22} to enhance the model capability, and then by the Factify 2 dataset.
In the feature extractor module, all images were transformed by resizing to 256, center cropping to 256, and normalizing.
On the other hand, all text was normalized by replacing emojis in text strings using demoji \footnote{https://pypi.org/project/demoji/}, all the abbreviations are expanded, and we delete the @name and URL to shorten the length of the text with meaningless words.
All the experiments were conducted on a machine with AMD Ryzen Threadripper 3960X 24-Core Processor, Nvidia GeForce RTX 3090, and 252GB RAM.
To evaluate the performance of the task, the weighted average F1 score was used across the 5 categories.
The source code is available at https://github.com/wwweiwei/Pre-CoFactv2-AAAI-2023.

\begin{table}[t]
  \centering


\begin{tabular}{c||c|ccccc}
    \toprule
    Module Used & Pre-CoFact & (1) & (2) & (3) & (4) & (5) \\
    \midrule
    Feature Extractor & x & v & v & x & v & x \\
    \midrule
    PLM\textsubscript{CV} & DT-b & CN & SW-b & SW-b & SW-b & SW-b \\
    \midrule
    PLM\textsubscript{NLP} & DE-b & BB & DE-l & DE-l & DE-l & DE-l \\
    \midrule
    Adapters & x & x & x & x & v & v \\
    \midrule
    SupConLoss & x & x & x & x & v & x \\
    \midrule
    \midrule
    Weighted F1 (\%) & 74.22 & 74.67 & 76.60 & 75.89 & 75.56 & \textbf{78.80} \\
    \bottomrule
\end{tabular}
  \caption{Variations of our model by validation score. The pre-trained models include: DeiT-base (DT-b), CoatNet (CN), Swinv2-base (SW-b), Deberta-base (DE-b), Deberta-large (DE-l), BigBird (BB).}
  \label{tab:experiment-ablation}
\end{table}

\subsection{Ablation Study}
\subsubsection{Variations of Our Proposed Model}
\label{variations}
To examine the relative contribution of our proposed module, we first conduct the ablative experiments by removing each module as the variants of Pre-CoFactv2.
Table \ref{tab:experiment-ablation} summarizes the results of the variations on the validation set.
We can observe that adopting different pre-trained models (i.e., CoatNet \cite{DBLP:conf/nips/DaiLLT21} and BigBird \cite{DBLP:conf/nips/ZaheerGDAAOPRWY20}) with feature extractor (1) slightly outperforms Pre-CoFact.
Moreover, using a pre-trained CV model with Swinv2 with an adapter and a pre-trained NLP model with Deberta-large (3) significantly improves the performance compared with Pre-CoFact, which illustrates the importance of not only larger PLMs and finetuning them with lightening parameters but also the multi-modal multi-type fusion. 
Comparing the variations (2) with (3), additional information from the feature extractor further boosts the performance by capturing explicit information instead of only semantic information from the embeddings.
It is worth noting that adding contrastive loss to the embeddings after fusion fails to benefit overall performance, which suggests that calculating the contrastive loss of images and text respectively may be more effective.
Nonetheless, our proposed framework signifies the ability to incorporate multi-modal claims and documents with explicit features and lightening parameters to effectively classify the veracity of the news.

\begin{table}
  \centering
\begin{tabular}{c|c|c|c|c}
    \toprule
    Model & (1) Average & (2) Weighted & (3) Power & (4) Unified power \\
     & & sum & weighted sum & weighted sum (Our) \\
    \midrule
    Support Text F1 (\%) & 71.64 & 72.65 & \textbf{73.50} & 73.46\\
    \midrule
    Support Multimodal F1 (\%) & 80.00 & 81.39 & 81.66  & \textbf{81.88}\\
    \midrule
    Insufficient Text F1 (\%) & 75.21 & 75.78 & \textbf{76.39} & 76.30\\
    \midrule
    Insufficient Multimodal F1 (\%) & 74.62 & 76.22 & 76.11  & \textbf{76.41}\\
    \midrule
    Refute F1 (\%) & 99.73 & 99.70 & \textbf{99.80} & \textbf{99.80}\\
    \midrule
    \midrule
    Weighted F1 (\%) & 80.24 & 81.15 & 81.49  & \textbf{81.57}\\
    \bottomrule
\end{tabular}
  \caption{Ablation study of ensemble techniques in terms of validation score. (1) Average: (prob1+prob2+prob3)/3, (2) Weighted sum: $w1\times prob1 + w2\times prob2 + w3\times prob3$, (3) Power weighted sum: $w1\times prob1^p + w2\times prob2^p + w3\times prob3^p$, (4) Unified power weighted sum (Ours): $w1\times prob1^{p1}+w2\times prob2^{p2}+w3\times prob3^{p3}$.}
  \label{tab:experiment-ensemble}
\end{table}

\subsubsection{Variations of Our Ensemble Technique}
To ensure the effectiveness of our unified ensemble technique, we conducted a comprehensive ablation study of variants of ensemble techniques by integrating all the predictions in different weights and powers.
(1) is the subset of (2) when all the weight is the same, (2) is the subset of (3) when all the power equals 1, and (3) is the subset of (4) when all the power is equal.
Table \ref{tab:experiment-ensemble} proves that the unified power weighted sum achieves the best result, which demonstrates that our proposed unified ensemble method is superior to other techniques by up to 1.7\%.

We can observe that averaging all predictions improves the performance only slightly, while setting different importance and powers boosts the performance more.
Our unified ensemble technique, in contrast, achieves the best quality compared to different ensemble methods.
We note however that these parameters are manually tuned, which requires more costs to achieve better performance.

\subsection{Testing Performance}
The results for the testing set are shown in Table \ref{tab:experiment-test} in terms of the weighted F1-score. 
Our approach achieved the state-of-the-art performance of 81.82\% of the weighted F1-score, winning first place in the multi-modal fact-checking challenge, and outperformed the second place and the official baseline by 1.3\% and 25.9\%, respectively.
This again indicates the reasonable and effective design of Pre-CoFactv2.

\begin{figure*}
  \centering
  \includegraphics[width=\linewidth]{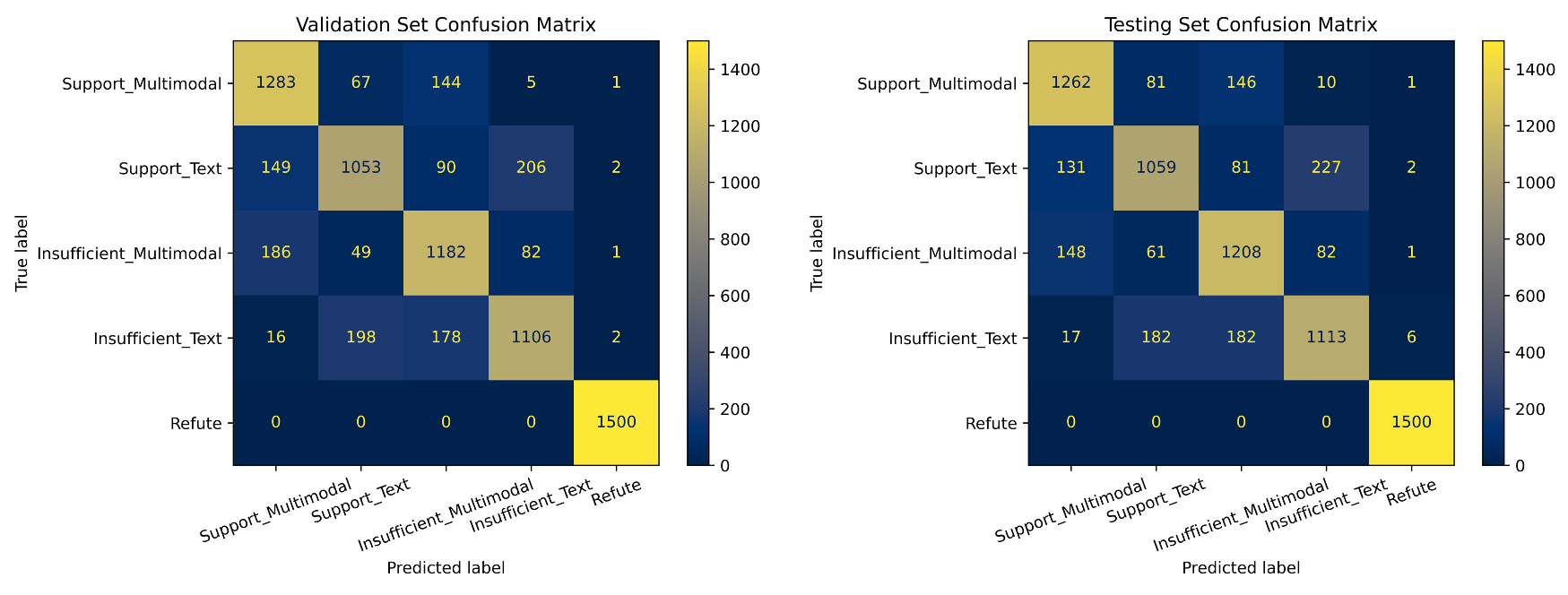}
  \caption{The confusion matrices of the validation set and testing set of our proposed model Pre-CoFactv2.}
  \label{fig:confusion-matrix}
\end{figure*}

To further analyze the classification details, we illustrate the confusion matrices of the validation set as well as the testing set of Pre-CoFactv2.
As shown in Figure \ref{fig:confusion-matrix}, we can observe that the class of refute is the most distinguishable category, while some of the classes between insufficient\_text and support\_text misclassify each other.

\begin{table}
  \small
  \centering
  \addtolength{\tabcolsep}{-2pt}

\begin{tabular}{c||cc}
\toprule
\multicolumn{1}{l}{}          & \textbf{Triple-Check} & Baseline \\ 
\midrule
Support\_Text (\%)            & \textbf{82.77}        & 50.00    \\
Support\_Multimodal (\%)      & \textbf{91.38}        & 82.72    \\
Insufficient\_Text (\%)       & \textbf{85.19}        & 80.24    \\
Insufficient\_Multimodal (\%) & \textbf{89.22}        & 75.93    \\
Refute (\%)                   & \textbf{100}          & 98.82    \\
\midrule
Final (\%)                    & \textbf{81.82}        & 64.99    \\ 
\bottomrule
\end{tabular}
  \caption{Performance of Pre-CoFactv2 in terms of testing score, which achieved the first place and outperformed the official baseline \cite{surya2023factifyoverview} by 25.9\%.}
  \label{tab:experiment-test}
\end{table}

\section{Conclusion}
In this work, we introduce parameter-efficient large foundation models with feature representation (Pre-CoFactv2) to mitigate the issue of disseminating multi-modal fake news.
With the integration of adapters with foundation models, we are able to finetune the backbones with only a few parameters while achieving better performance.
In addition, feature representations provide explicit information about both text and images of claims as well as documents to clarify the relations between them with multi-modal multi-type fusion.
With the help of a unified ensemble technique, Pre-CoFactv2 is ranked first on the official leaderboard with an F1 score of 0.81, which outperforms the baseline by 25.9\% and greatly benefits the research of multi-modal fact verification.
Furthermore, extensive ablative experiments demonstrate the effectiveness of each module in our proposed framework.

\bibliography{reference}

\end{document}